# Improving Patient Pre-screening for Clinical Trials: Assisting Physicians with Large Language Models

**Danny M. den Hamer**[1, ✉, *], **Perry Schoor**[1,*], **Tobias B. Polak**[1-3], **Daniel Kapitan**[4]

[1] myTomorrows, Amsterdam, the Netherlands.
[2] Department of Biostatistics and Epidemiology, Erasmus MC, Rotterdam, The Netherlands
[3] Erasmus School of Health Policy and Management, Erasmus University Rotterdam, Rotterdam, The Netherlands
[4] Eindhoven University of Technology, Department of Mathematics and Computer Science, Eindhoven, the Netherlands.
* These authors contributed equally.
✉ email: danny.den.hamer@mytomorrows.com

ORCID-ID:
T.B. Polak:        0000-0002-2720-3479
D. Kapitan:        0000-0001-8979-9194

## Abstract

**Background** Physicians considering clinical trials for their patients are met with the laborious process of checking many text based eligibility criteria. Large Language Models (LLMs) have shown to perform well for clinical information extraction and clinical reasoning, including medical tests, but not yet in real-world scenarios. This paper investigates the use of InstructGPT to assist physicians in determining eligibility for clinical trials based on a patient's summarised medical profile.

**Methods** Using a prompting strategy combining one-shot, selection-inference and chain-of-thought techniques, we investigate the performance of LLMs on 10 synthetically created patient profiles. Performance is evaluated at four levels: ability to identify screenable eligibility criteria from a trial given a medical profile; ability to classify for each individual criterion whether the patient qualifies; the overall classification whether a patient is eligible for a clinical trial and the percentage of criteria to be screened by physician.

**Findings** We evaluated against 146 clinical trials and a total of 4,135 eligibility criteria. The LLM was able to correctly identify the screenability of 72% (2,994/4,135) of the criteria. Additionally, 72% (341/471) of the screenable criteria were evaluated correctly. The resulting trial level classification as eligible or ineligible resulted in a recall of 0.5. By leveraging LLMs with a physician-in-the-loop, a recall of 1.0 and precision of 0.71 on clinical trial level can be achieved while reducing the amount of criteria to be checked by an estimated 90%.

**Interpretation** LLMs can be used to assist physicians with pre-screening of patients for clinical trials. By forcing instruction-tuned LLMs to produce chain-of-thought responses, the reasoning can be made transparent to and the decision process becomes amenable by physicians, thereby making such a system feasible for use in real-world scenarios.

**Funding** myTomorrows.

# Introduction

Physicians considering clinical trials for their patients are often faced with a large amount of options which are hard to navigate. The overall process of enrolling patients in clinical trials typically involves three steps: i) finding suitable trials in clinical trial registries through filtering on patient demographics like age, sex, country and medical condition; ii) pre-screening of trial eligibility criteria, by comparing the patient's medical profile to the trial's eligibility criteria; and iii) selecting the appropriate trial. It is widely recognized that a major obstacle is the arduous pre-screening process, during which hospital staff examine patients' medical histories to the trial criteria.[1] One potential solution to streamline this process is to explore automated eligibility pre-screening. In this regard, Natural Language Processing (NLP) is naturally considered since the eligibility criteria are unstructured text.

The majority of current approaches to automated pre-screening of clinical trial matching are based on NLP pipelines that use information extraction through sentence segmentation, named-entity recognition (NER) and relation detection, combined with downstream normalisation, rule-based processing and querying. Idnay et al. reviewed 11 such NLP-based systems which all evaluated the capacity of the system to identify patients as eligible.[2] However, no standardised evaluation test exists, resulting in various performance metrics and widely varying amounts of patients or trials tested, making it difficult to coherently summarise the current state-of-the-art performance.[1] Of the 11 studies, 7 were reported to specifically focus on increasing workload efficiency in a 'hybrid intelligence' setting where the AI system is used for pre-screening whilst a physician-in-the-loop is responsible for making the final assessments. Again, no standardised evaluation was used. Common among the studies reviewed was the limited number of trials evaluated, each covering less than 50 trials.[1] Criteria2Query 2.0, a more recent system published after the review by Idnay et. al.[2], managed to achieve an end-to-end accuracy on trial level of approximately 50%. This was achieved on a much larger number of clinical trials (1,010), however, it was achieved in COVID-19 as a singular disease area. It is unclear how this performance translates to other disease areas.[3,4]

Clearly, there is still much room for improvement in the automation of eligibility pre-screening, especially through the use of NLP. First results indicate that Large Language Models (LLMs), an emerging subtype of NLP, can significantly outperform conventional NLP systems for a variety of tasks.[5] An insurgence of studies is ongoing to demonstrate the performance of LLMs (for example, GPT, PaLM, LaMDA) in combination with their instruction-tuned variant with reinforced learning with human feedback (InstructGPT, Flan-PaLM, Bard, respectively) in extracting clinical information, encoding clinical knowledge and performing clinical reasoning tasks.[6–9]

At myTomorrows, a service provider for patients and physicians looking for treatment options, we receive many medical profiles which are subsequently screened by medical professionals to match to clinical trials. We wondered whether the workload of this process, which is partly manual and thereby resource and time intensive, could be reduced, whilst ensuring medical oversight. Combining the experience from conventional NLP-based systems with the preliminary findings on the potential of LLMs, this paper reports the design, use and performance of an LLM-assisted pre-screening of eligibility criteria of clinical trials using OpenAI's InstructGPT with text-davinci-003. The approach taken here aims to improve workload efficiency with a physician-in-the-loop, focusing on workload efficiency, whilst at the same time taking a patient-centric approach on pre-screening, prioritising high recall over time efficiency.

# Methods

Our simulation setup aims to resemble the real-life situation of physicians and patients that are looking for investigational treatment options. Based on the basic patient information, including primary disease, age, gender, and geographical location, physicians search international databases such as clinicaltrials.gov to find a list of potential trials. Subsequently, the detailed patient's medical background is carefully assessed to match all the individual eligibility criteria of all retrieved trials. A subset of eligible trials is selected for which patients meet all criteria.

To address the limitations of conventional NLP-based systems for pre-screening of eligibility criteria, we propose a methodology that leverages the medical reasoning capacity of instruction-tuned LLMs. The system architecture and data flow of Criteria2Query 2.0 is taken as a reference, that is, the task of eligibility screening is conceived as a combination of information retrieval and reasoning.[3] Specifically, we instruct the LLM to reason what criteria can be screened and whether the medical profile meets these criteria.

Our setup can be ordered as follows:
1. Creating synthetic patient profiles as input
2. Collecting relevant clinical trials
3. Retrieving and preprocessing their eligibility criteria
4. Identifying which criteria can be screened based on the information provided
5. Evaluate whether the patient meets these criteria based on the information provided
6. Determining the patient's eligibility for the trials based on unmet criteria

We additionally measure the percentage of criteria to be checked by a physician-in-the-loop based on the statistics regarding the setup mentioned above.

**Synthetic patient profiles**
To preserve patient privacy, we employ synthetic rather than real-world patient profiles. These patient profiles consist of two parts: basic demographic information (medical condition, country of residence, age and biological sex) and the detailed medical summary. The medical summary is a piece of unstructured text describing in-depth medical information relevant for pre-screening. We create 10 distinct patient profiles, each covering a different disease that is randomly selected from the real-world patient profiles we have received for screening over the last year.

Based on the experience of screening thousands of patients, certain categories in the eligibility criteria were identified as distinguishably common categories, specifically: diagnosis and histology, previous treatment, lab values and test results, biomarkers, concurrent disease and overall health (e.g. ECOG-score, life expectancy). Based on these common categories, medical professionals created the summaries. It is noteworthy that the profiles may contain acronyms, spelling errors, and incomplete information, to mimic real-world situations. The synthetic patient profiles are included in the supplemental material. An example profile is included in figure 1.

**Collecting relevant trials**
Following this, the 10 artificial patient profiles, each corresponding to a disease represented by a Medical Subject Headings (MeSH) code, are employed to conduct a search for interventional clinical trials using myTomorrows' Treatment Search, which combines well-known trial registries clinicaltrials.gov, and EudraCT. Subsequently, we extract and process the unstructured eligibility criteria text for all the identified trials to ensure machine readability; this processing entails dividing the full text into two sections, inclusion and exclusion, after which these sections are split into individual criteria. Clinical trials featuring unstructured data that is unsuitable for automated text processing, such as trials without a distinct separation between inclusion and exclusion criteria, will be omitted from the assessment. In practice, these trials are directly sent to the physician-in-the-loop for manual evaluation.

## Prompting strategy: assessing individual eligibility criteria

The prompting strategy is based on emerging best practices and is illustrated in figure 1. To identify what criteria are screenable based on the medical summary, we use a selection-inference prompt.[10] Subsequently, to cause the LLM to reason whether the profile meets each criterion, we employ chain-of-thought prompting techniques.[8] For each selected criterion the LLM is prompted to reason why it is met, not met or unknown. The 'unknown' category is included to allow the LLM to indicate when it is uncertain and/or cannot provide an answer based on given the input, to prevent hallucinations.[11] Finally, to allow automatic processing of the outputs, the model is prompted to convert the reasoning per criterion to a machine readable value ('met', 'not met' or 'unknown'). To provide the LLM an example of the expected behaviour we additionally employ a one-shot approach, as this generally increases LLM performance.[12] All experiments were conducted on InstructGPT using OpenAI's model text-davinci-003, and performed at temperature 0 to improve reproducibility of the experiments.

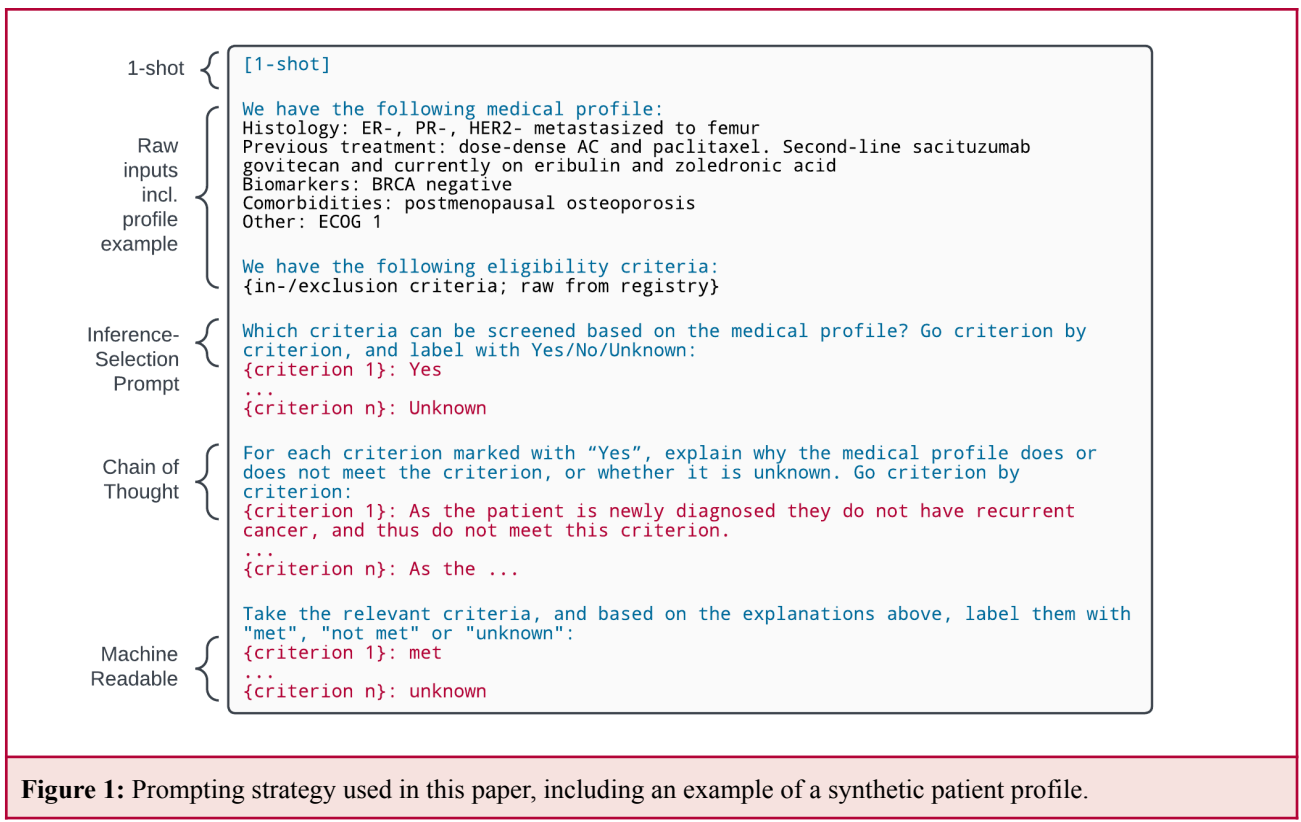

**Figure 1:** Prompting strategy used in this paper, including an example of a synthetic patient profile.

## Performance metrics

### Processing individual eligibility criteria and physician evaluation

To assess individual eligibility criteria, we measure three performance metrics at the criterion level.

1. Identification of criteria as screenable: We consider a criterion screenable if the medical profile contains sufficient information to evaluate the criterion. We assume all relevant information is provided by the physician in the medical summary (closed-world assumption), with the exception of protein expression and gene mutations. For criteria unrelated to the studied disease, the open-world assumption is applied, where missing information cannot be assumed absent.
2. Medical reasoning: classifying reasoning errors using categories presented by Liévin et. al. (Pattern D - Incorrect reasoning step, Pattern E - Incorrect/insufficient knowledge, and Pattern F - Incorrect reading).[9] This helps evaluate the LLM's chain-of-thought reasoning.

3. Evaluation of individual criterion eligibility: As a subset of all criteria we define 'dropout criteria', criteria which cause trials to be considered ineligible (not-met inclusion or met exclusion criteria). Dropout criteria impact recall at the trial level.

To ensure no eligible trial is missed, a physician-in-the-loop reviews all dropout criteria. We evaluate the quantity of dropout criteria reviewed by the physician due to the LLM's output, comparing it to the actual amount of dropout criteria in the dataset and the amount required during manual pre-screening, which we assume is 100%.

**Evaluating trial eligibility**
The primary goal of a physician is to provide patients a comprehensive overview of all treatment options available. This involves the physician reviewing trials and their eligibility criteria and categorising clinical trials as either eligible or ineligible based on the patient's medical profile. The critical metric to optimise in this process is the recall on the trial level, which entails identifying all eligible trials for the patient. Focusing on recall ensures that no potential treatment options are overlooked during the decision-making process.

As there is no gold standard for evaluating performance at the recall and precision level, a licensed medical professional manually reviews the LLM's output for automatic pre-screening according to the definitions laid out above.[3] To investigate the system's overall performance and generalizability, we focus on the collective results instead of individual patient performance. Detailed results and manual evaluations are available in the supplemental material.

**Evaluating stochasticity of outputs**
To obtain a first indication of stochasticity of the generated output, all patient profiles have been run 10 times against the same set of selected clinical trials, at two different temperature settings (0, 1) using text-davinci-003. A single run on GPT-4 was performed for comparison.

# Results
Table 1 shows an overview of the 10 synthetic patient profiles. The queries were performed between February 8 and February 13, 2023. The search queries for these profiles resulted in 146 trials containing 4,135 criteria.

| Profile | Disease | MeSH code | Trials (n) | Criteria (n) |
|---|---|---|---|---|
| FP001 | Cancer of the Uterine Cervix | M0003943 | 10 | 302 |
| FP002 | Leukemia, Myeloid, Acute | M0023827 | 11 | 335 |
| FP003 | Colorectal Cancer | M000655646 | 15 | 489 |
| FP004 | Cholangiocarcinoma | M0027512 | 34 | 1,109 |
| FP005 | Glioblastoma | M0009269 | 9 | 252 |
| FP006 | Muscular Dystrophy, Duchenne | M0014253 | 10 | 158 |
| FP007 | Breast Cancer | M0002909 | 17 | 420 |
| FP008 | Amyotrophic Lateral Sclerosis | M0001056 | 5 | 103 |
| FP009 | Cancer of Pancreas | M0333232 | 14 | 402 |
| FP010 | Carcinoma, Non-Small-Cell Lung | M0003440 | 21 | 565 |
| | | **Total** | **146** | **4,135** |

**Table 1:** List of synthetic profiles, associated diseases, number of trials and number of criteria included in this report. Medical Subject Headings (MeSH) codes are internationally standardised disease codes.

**Evaluation at criterion level**

Reasoning performance is described in table 2. The model's capacity to correctly identify criteria as screenable and ability to correctly identify dropout criteria is illustrated in figure 2. A comprehensive overview of all results, including criterion evaluation capacity for non-dropout criteria, is available in the supplemental material.

| Is screenable | Type of error | Errors (n) |
|---|---|---|
| TP | Incorrect reasoning (D) | 85 |
| TP | Incorrect or insufficient knowledge (E) | 13 |
| TP | Incorrect reading (F) | 38 |
| FP | Incorrect reasoning (D) | 442 |

**Table 2:** count of type of errors in evaluating criteria

The model's performance at correctly identifying criteria as screenable resulted in an accuracy of 72% (2,994/4,135). Most noteworthy to discuss here is the difference in performance of both medical reasoning and evaluating individual criteria, between correctly and incorrectly identified as screenable criteria. Within 471 screenable and screened criteria (TP), 132 (28%) reasoning errors occurred. Focusing just on dropout criteria, 410 (87%) were correctly classified with 28 false negative dropout criteria and 33 false positive dropout criteria. Within the 865 not screenable but screened (FP) dropout criteria, 442 (51%) reasoning errors occurred. All reasoning errors were automatically classified as Pattern D, as the model incorrectly reasoned towards a conclusion while not enough information was present. These FP criteria also resulted in 187 false positive dropout criteria. Comparing TP to FP, it is clear the majority of false positive criteria result from not screenable but screened criteria (33 vs 187 respectively), resulting in this category being the main driver for lowering recall.

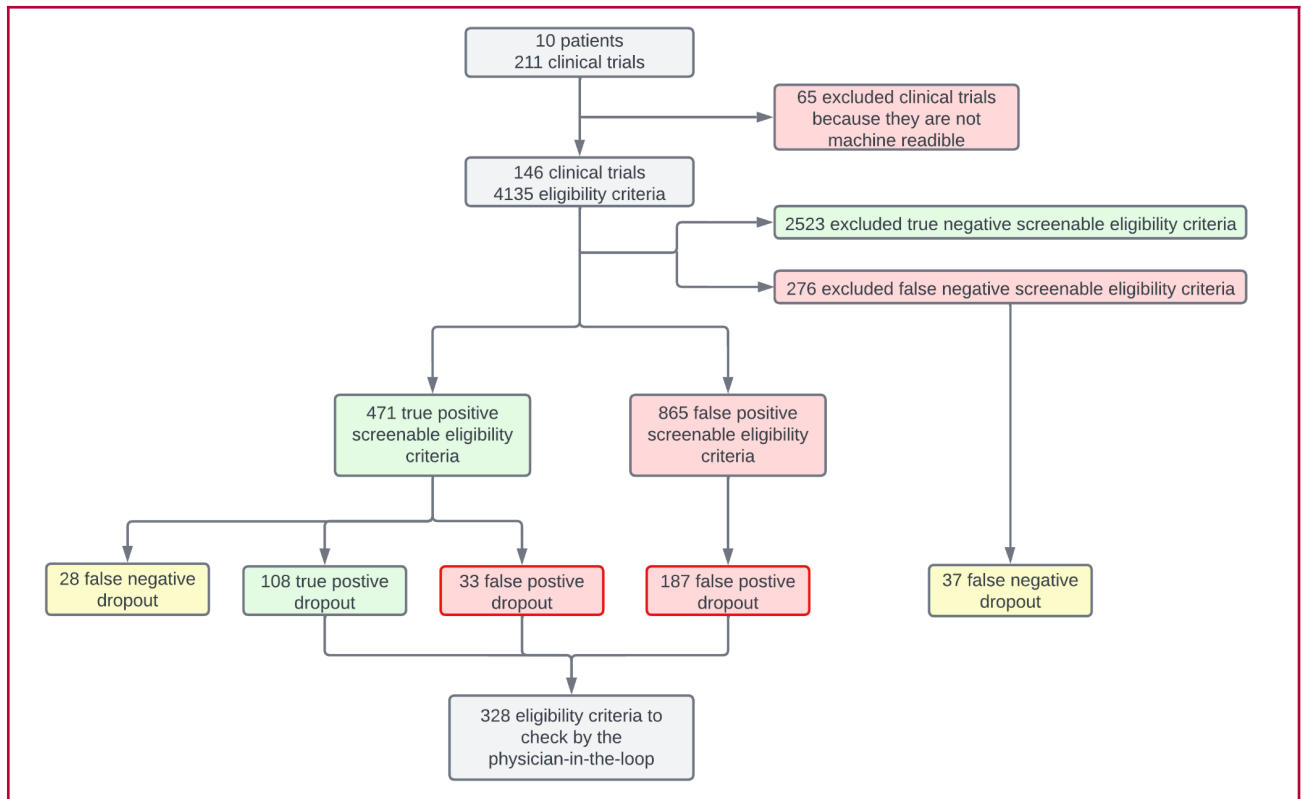

**Figure 2**: Workflow for an efficient evaluation of eligibility criteria for selecting suitable clinical trials for a patient by a physician-in-the-loop. The total amount of eligibility criteria to be checked is reduced from 4135 to 328. The numbers outlined in red are the amount of eligibility criteria that unnecessarily need to be checked by the physician; this will increase time to be invested by the physician. The numbers highlighted in yellow are the amount of criteria that are incorrectly put in the final overview of clinical trials; these decrease precision.

### Evaluation at trial level

Table 3 shows the confusion matrix of the evaluation at trial level. Out of 146 evaluated trials, 68% were accurately classified as being eligible to the patient (TP 32, 22%) or ineligible (TN 68, 47%), resulting in an overall precision of 0.71. With 33 false negatives (23% of total), the recall at trial level is 0.50. The largest factor resulting in a lower recall are the 220 false positive dropout criteria. Of these, 85% are caused by false positive "is screenable" in step 2 of the workflow.

|  | **Actual eligible** | **Actual ineligible** |
| --- | --- | --- |
| **Predicted eligible** | 32 (TP) | 13 (FP) |
| **Predicted ineligible** | 33 (FN) | 68 (TN) |
| Trials not evaluated due to automatic processing errors: 36 | | |

**Table 3:** Confusion matrix of the 146 trials as listed in table 1 that were evaluated with the system. An additional 36 trials were returned from initial search, but were not evaluated with the system due to processing errors.

### Impact on physician evaluation

In practice, the final step in the workflow involves the evaluation by a physician-in-the-loop which is envisaged as follows. First, all criteria that have been classified as a dropout criterion are evaluated. This includes both TP and FP dropouts, 328 criteria in total as seen in figure 2, the majority of which originates from not screenable but screened criteria. Assuming the physician-in-the-loop is able to correct all false positive dropout criteria, this processing step effectively establishes a recall of 1, resulting in an overall performance of the process to a recall of 1 with a precision of 0.71 as calculated in the subsequent paragraph. This constitutes a significant reduction in workload to less than 10% (328/4,135) of the total number of criteria need to be evaluated to achieve this performance.

**Stochasticity of output**

The results for 10 different runs on text-davinci-003 for step 1 (screenability of criteria) and step 3 (evaluation at trial level) are shown in figure 3. Full details of the results of these experiments are included in the supplemental material. We observe that the variability of the LLM indeed is lower at temperature 0, where results between runs have smaller variation in precision and recall compared to temperature 1. In the case of patient profile 6, for example, we measure a standard deviation of 0.033 (T=0) vs. 0.082 (T=1) for the precision of step 1 (screenability, top row). Similarly, the recall at trial level for patient 6 (bottom row) has a standard deviation of 0.022 (T=0) vs. 0.131 (T=1). We tentatively conclude that setting T=0 indeed improves the reproducibility of results. A single run on GPT-4 at temperature 0 with the same scripts yielded similar results as text-davinci-003 at temperature 0. Reliability and quantification of uncertainty will be discussed below.

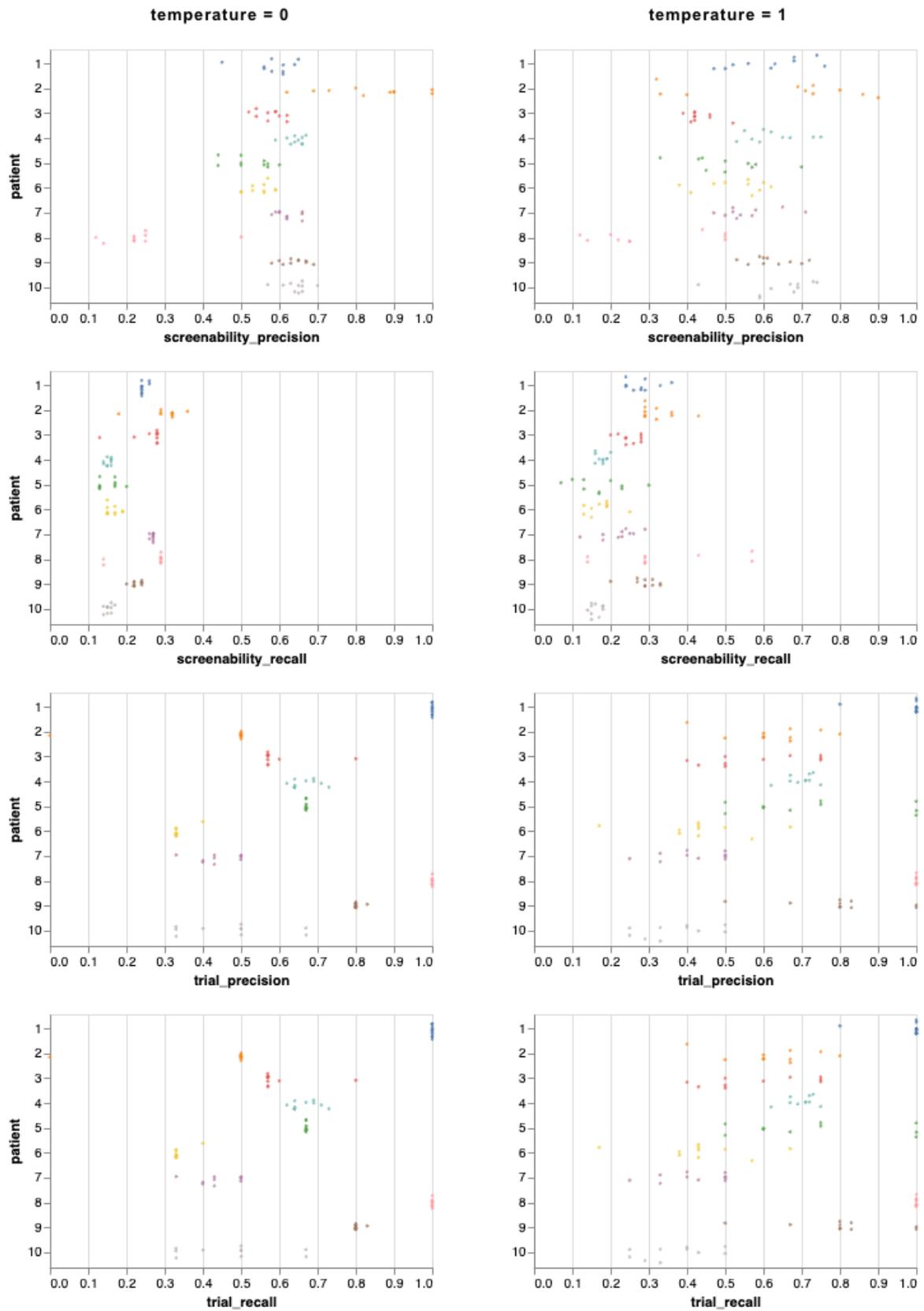

**Figure 4**: Precision and recall for 10 runs for each patient profile at temperature 0 (left column) and temperature 1 (right column). Each circle represents the precision and recall for one run, against the same set of eligibility criteria and clinical trials.

# Discussion

**Reduction of workload through pre-screening**

In this research, we have demonstrated LLMs have the potential to increase efficiency in trial pre-screening for medical professionals. Given an unstructured text medical summary as input, LLMs can (i) correctly identify if there is enough information in the medical summary to evaluate eligibility criteria in 72% (2994/4135) of the cases, (ii) answer eligibility criteria correctly in 72% (341/471) of the cases if adequate levels of information is present in the medical summary, (iii) significantly reduce physicians' workload, even if the optimal use of LLMs requires their close oversight.

Our work adds to the existing literature as a first attempt to utilise LLMs to partially automate clinical trial pre-screening for patients. This extends the prior focus on NLP-models. Compared with other studies, such as (part of the) the work described in Idnay et al. and Criteria2Query 2.0, our study covers a range of diseases rather than focusing on a single disease area, improving generalizability.[1,4] As a downside, we only evaluate our results for a limited number of patients, and increasing patient numbers may lead to more accurate evaluation of our method.

Our method has the potential to reduce physicians' workload, but we have identified specific peculiar behaviour, especially when the LLM is overconfident in analysing unclear individual trial criteria. If the LLM accurately determines that there is adequate information in the patient profile to evaluate individual eligibility criteria, our model allows to fully automate a considerable amount of manual work. However, the incorrect reasoning of the LLM in case of inadequate patient information is a cause of concern and impedes the full automation of trial pre-screening. The review of Idnay et al. describes a similar mediocre performance of automated medical classification models.[3] Our model falsely classifies 50% of potentially eligible clinical trials as ineligible. We counter this poor performance of misclassification by incorporating a physician-in-a-loop, who needs to approve suggestions by the LLM to classify a clinical trial as ineligible. Although this safeguard still requires manual work by the physician, we accomplished to reduce the amount of eligibility criteria to be checked by 90%.

Our method can be improved in several ways. For example, we have seen that 67% (220/328) of the eligibility criteria that need to be checked by the physician are actually not crucial in classifying a clinical trial as ineligible and 57% (187/328) is caused by the LLM falsely reasoning it has enough information to evaluate the given eligibility criteria. Therefore, the biggest improvement can be made by focusing on preventing hallucination at the stage where the LLM identifies criteria as screenable.

Hallucination is a frequently mentioned down-side of LLMs. Although the impact of hallucination may vary per use case, when it comes to assessing the best options for experimental treatments for patients, it is imperative to avoid unwanted LLM-behaviour. Keeping a physician-in-the-loop cannot only prevent the machine from hallucination, but may also be of use when it comes to interpreting trial results and answering questions from patients. Before such methods can be used in clinical practice, there remains a variety of ethical concerns to be addressed, including issues of responsibility, quality standards, and patient safety.

**Ethical use of LLMs for pre-screening of eligibility criteria**

The recent surge of work done on LLMs has also led to a fierce debate on the ethical aspects of using these systems in real-world scenarios. In fact, at the time of writing the Future of Life Institute called to pause

"giant AI experiments" for at least 6 months.[13] To address these concerns, Harrer S. considered how LLMs could be developed ethically within the framework put forward by the the World Health Organization.[14] We consider our findings in this context in table 4.

| Aspect | Evaluation |
| --- | --- |
| Accountability | We argue that the workflow proposed is supportive of leaving accountability with the users of the system. By exposing the reasoning/output by the LLM to the end user, and leaving responsibility of any decisions made to be mandatorily reviewed by the medical professional, we defer accountability to the treating physician. |
| Fairness | The work presented here uses models and APIs from OpenAI, which are known to be severely lacking in terms of providing the necessary information to address potential bias or misinformation. We argue, however, that the work presented here is of value to the healthcare and research community at large, and foresee reproduction of the approach using other instruction-tuned LLMs such as Anthropic's Claude[15] or Google's Med-PaLM.[7] |
| Data privacy and selection | We have no direct influence on what data is ingested into LLMs to render them useful for our use-case. Hence we defer this issue until there is more clarity on the legal aspects of using LLMs. |
| Transparency and explainability | As we do not have automated decision making by the LLM, but rather show "it's reasoning" to the end user, and this reasoning is only used to assist the user, the transparency of the "decision making process" of the LLM internally is not an issue in this use-case. Rather, by framing the information retrieval and reasoning task as a closed answering system, we can monitor the performance in detail. We foresee a standardised benchmarking approach for continuous development and testing of instruction-tuned LLMs for pre-screening that contains hundreds of synthetic patient profiles which are validated against human input. By publishing our methods and synthetic data in this paper, we intend to contribute to transparency and explainability of LLMs for these types of use-cases. |
| Value and purpose alignment | We argue that using LLMs to expedite pre-screening of patient eligibility for clinical trials contributes to the values as set out in the Hippocratic Oath. We foresee no increased risk of harm, by using the hybrid intelligence approach with the physician-in-the-loop. Moreover, the approach presented here has the potential to make trials accessible to more people by improving the mediation process between patients and studies. In our evaluation, the approach presented here would be categorised as a low-risk AI system within the nascent European AI Act.[16] |

**Table 4:** Ethical evaluation within the framework put forward by the the World Health Organization.[14]

## Limitations

Our work needs to be interpreted in light of several limitations. First, the current experiment did not employ the train-test validation strategy commonly used in developing machine learning systems. It may be the case that the prompt engineering strategy has caused overfitting of the strategy on the 10 patient profiles provided. Comparing results from different runs shows that there is stochastic variation in the outputs even at temperature 0. More insight into calibrating the output and quantification of uncertainty is needed.

A model makes calibrated predictions if the probability it assigns to outcomes coincides with the frequency with which these outcomes actually occur. Language models are known to produce calibrated token-level probabilities. The question, however, is whether the output of LLMs in a supervised multiple-choice question-answering task (as is used here) can be meaningfully calibrated and thus provide a means for uncertainty quantification. Kadavath et al. have shown that LLMs are well-calibrated on True/False tasks, from which they propose the concept of self-evaluation, where the LLM is asked to evaluate its first answer in a True/False task.[18] While they demonstrated promising results for short-form answer tasks, this approach does not seem viable for chain-of-thought and other long-form tasks. In addition, given the high class imbalance of pre-screening, it is questionable whether this approach is viable for our use-case.

Kumar et al. have adapted conformal prediction for MCQA tasks to provide distribution-free uncertainty quantification.[17] Conformal calibration does not require any assumptions of the distribution of possible outcomes. Instead, it relies on computing a coverage guarantee from a known set of possible answers. They have shown that the uncertainty estimates from conformal prediction are tightly correlated with prediction accuracy of multiple-choice question-answering. We believe this approach is the most appropriate for the work presented here. This requires more detailed inspection of prompting strategies and experimenting with the log-probabilities that LLMs return for each generated token, which we defer to future work.

Second, our approach relies on the quality of the input data, including the ability to isolate every eligibility criterion to let the LLM reason on it and a clear distinction between inclusion and exclusion criteria. If the quality of the output of such parsing and isolating eligibility criteria cannot be guaranteed, the eligibility criteria of that trial cannot be processed by the system. In the experiments reported here, this amounted to around 30% of all the trials retrieved from the initial search, in line with earlier work as reviewed by Idnay et. al.[2] Given our design principle of optimising recall, trials that cannot be preprocessed correctly by the system need to be manually checked by the physician. This workload is added to the checking of the dropout criteria. Hence, 90% reduction in eligibility criteria that need to be checked cannot be directly translated into 90% reduction of workload for the physician.

Third, our approach was not evaluated on a standardised evaluation dataset as at the time of writing such a dataset does not exist.[1] We intend to contribute here by sharing our evaluation dataset which is unidentifiable and has been constructed by licensed medical professionals, available in the supplemental materials.

Finally, 43 of the 4135 eligibility criteria were incorrectly not classified as dropout criteria. As a consequence, clinical trials containing those false negative dropout criteria could be falsely classified as eligible, hampering precision of the final overview of suitable clinical trials based on the medical summary.

**Future work**
We expect that instruction-tuned LLMs can be used reliably assisting in pre-screening of eligibility criteria. At the time of writing, GPT-4 and other, more advanced instruction-tuned LLMs were announced. On top of that, a lot of medically focused LLMs are underway, including Med-PaLM[7], BioGPT[5] and Med-PaLM 2.[19] We foresee that the approach described here can be developed and optimised further with these improved LLMs. Just by implementing GPT-4 would probably already increase performance in both the screenability identification task as well as improving the correct answer percentage.[20] Furthermore, different prompting strategies can be explored. In the case of GPT-4, one strategy could be to ask the system to review its own output.[21] This strategy could be very beneficial on the "identify as screenable" step, as this is where the largest portion of false positive dropout criteria are sourced from.

Moving forward, the logical next step would be to evaluate the performance of the model on real patient data with inputs provided by healthcare professionals. Uncertainty quantification of the multiple-choice question answering can be done through conformal prediction. This will help assess the effectiveness and reliability of the tool in a practical setting. One major focus of this research should be to analyse how often healthcare professionals fail to identify false positive dropout criteria. Identifying and addressing this issue can significantly improve the accuracy and reliability of the model, making it a valuable tool that can be used in an ethical manner in healthcare to reduce 90% of the workload.

## Conclusion

To our knowledge, we are the first to introduce LLMs to assist physicians in eligibility pre-screening for clinical trials. By combining LLMs with a physician-in-the-loop, the described approach can reduce 90% of the eligibility criteria to be checked. By forcing instruction-tuned LLMs to produce chain-of-thought responses, the reasoning can be made transparent, and the decision process becomes amenable to physicians. This approach could make such a system practical, allowing LLMs to safely assist with finding and pre-screening clinical trials for patients, although real-world evaluation is needed.

## Contributors

All authors contributed equally to the conceptualization of the study and development of the methodology. DMdH and PS conducted the data preparation, curation and LLM experiments. DK drafted the manuscript, while TBP, DMdH, and PS critically revised the work. All authors approved the final version of this manuscript.

## Declaration of interests

DMdH, PS, and TBP report receiving personal fees from and owning stock/stock options (<0.1%) in myTomorrows; TBP reports receiving grants from HealthHolland and Prins Bernhard Cultuurfonds and being an unpaid member of the New York University Grossmann School of Medicine Ethics and Real-World Evidence Working Group. DK reports receiving funding from myTomorrows to contribute to this work as a contract researcher.

## Data sharing and supplemental materials

The synthetic patient profiles, tables with results and manual evaluations are available at
https://github.com/mytomorrows/llm-prescreening.